\documentclass{article}

\usepackage{arxiv}
\usepackage{fancyhdr} 
\usepackage{amsmath}   
\usepackage{amssymb}   
\usepackage{geometry}  

\usepackage{caption}

\usepackage[utf8]{inputenc} 
\usepackage[T1]{fontenc}    
\usepackage{hyperref}       
\usepackage{url}            
\usepackage{booktabs}       
\usepackage{amsfonts}       
\usepackage{nicefrac}       
\usepackage{microtype}      
\usepackage{lipsum}
\usepackage{graphicx}
\graphicspath{ {./images/} }
\usepackage{scalerel} 

\newcommand{\emojiGrinningFaceWithBigEyes}{\scalerel{\includegraphics{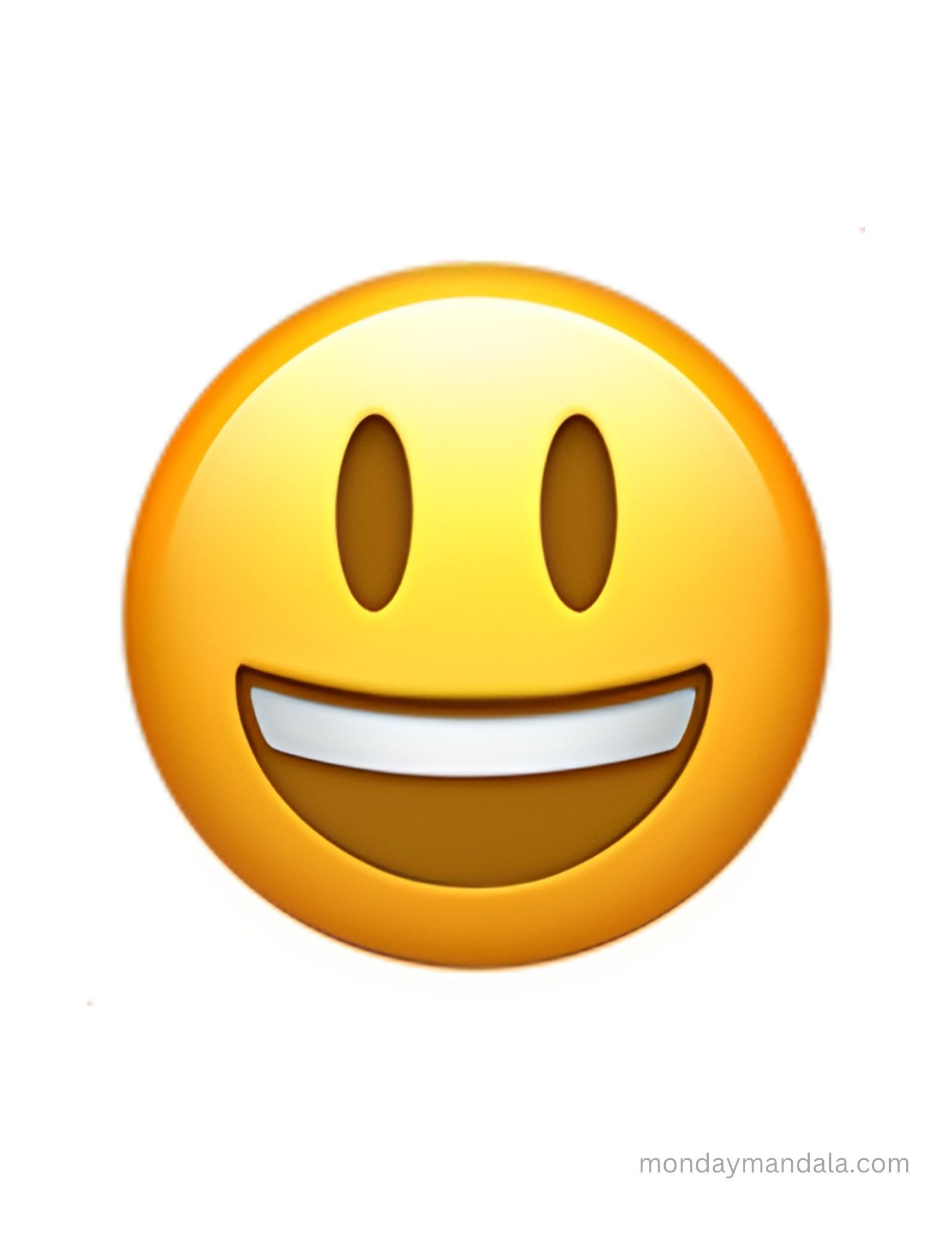}}{\strut}}
\newcommand{\emojiGrinningSquintingFace}{\scalerel{\includegraphics{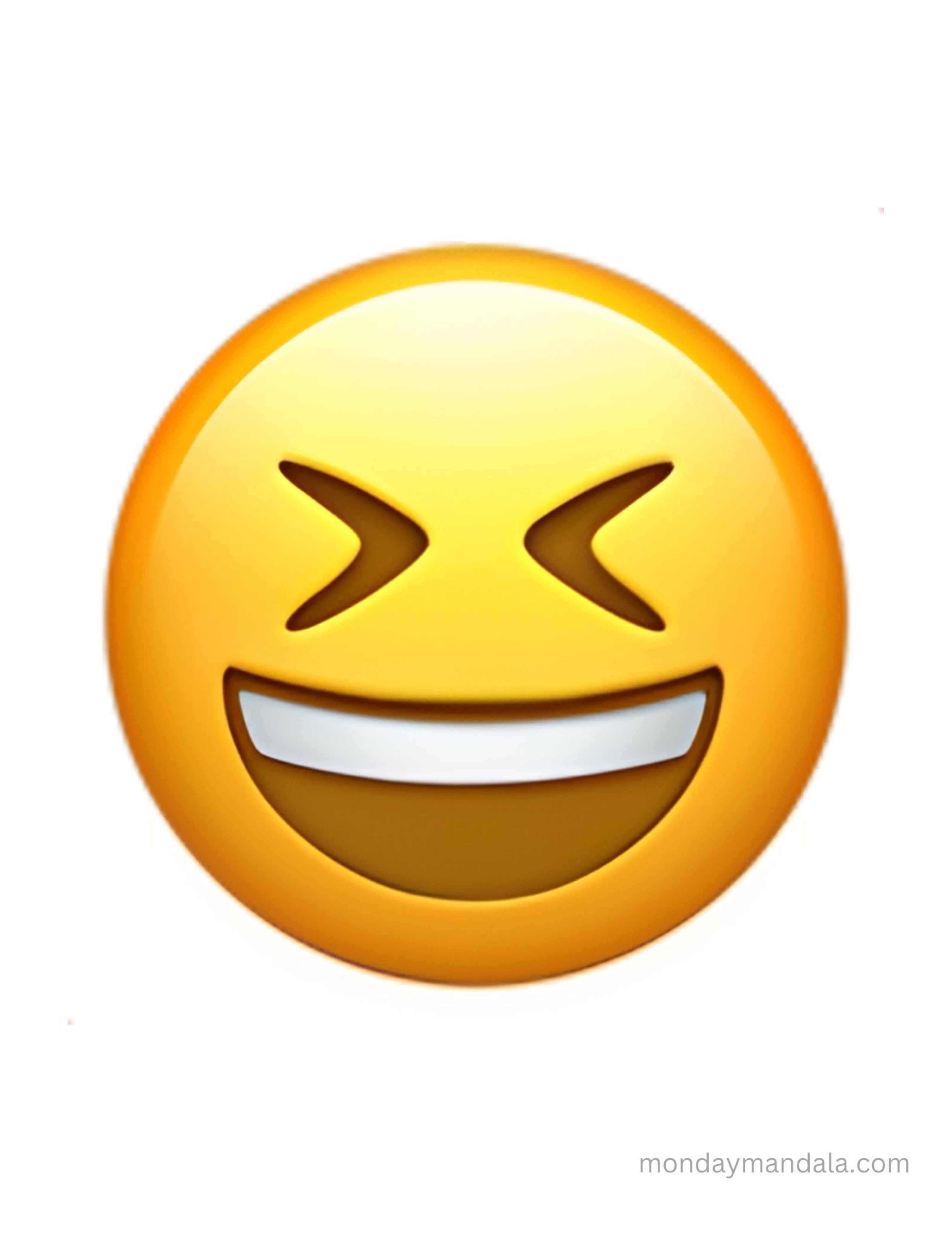}}{\strut}}
\newcommand{\emojiGrinningFaceWithSmilingEyes}{\scalerel{\includegraphics{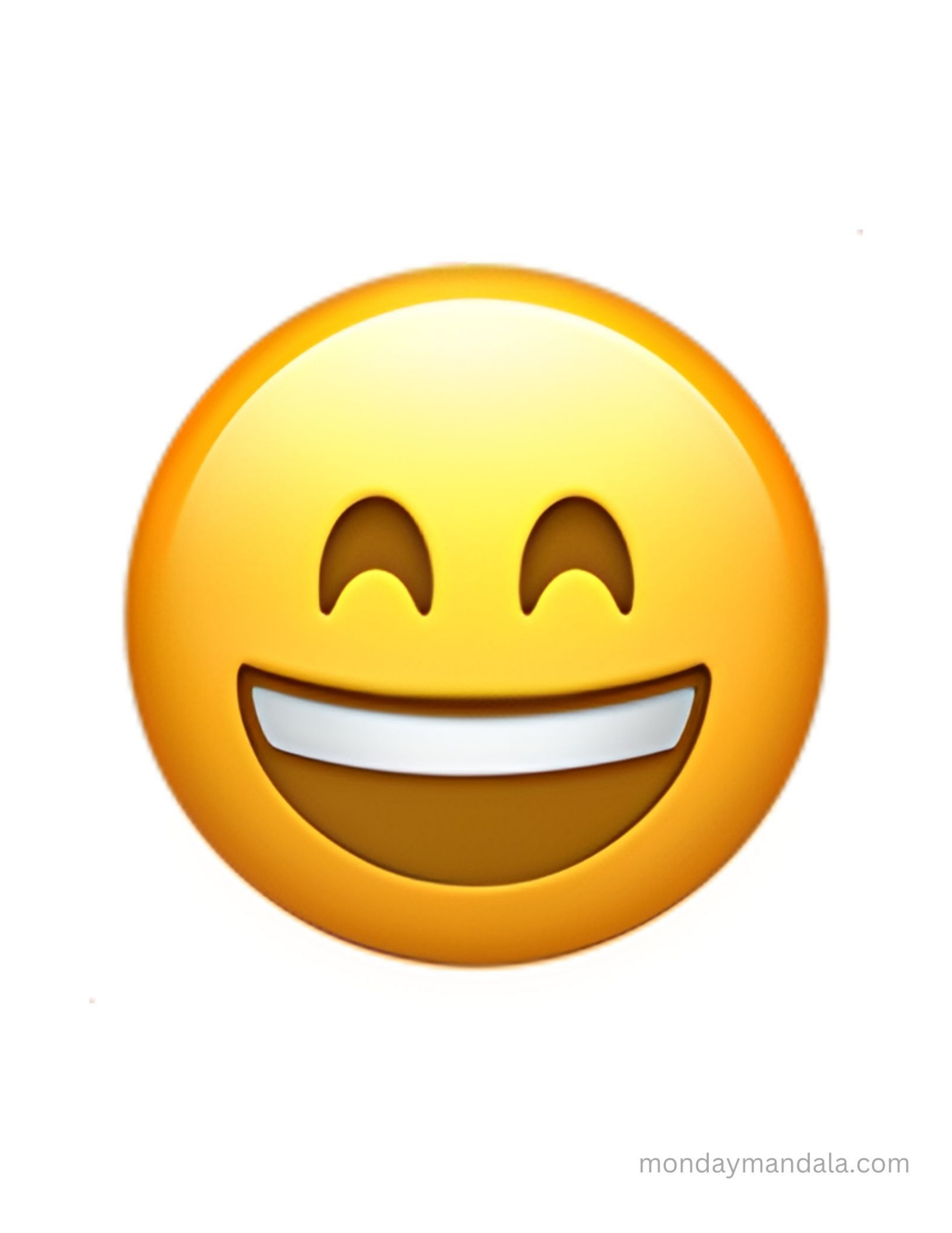}}{\strut}}
\newcommand{\emojiGrinningFaceWithSweat}{\scalerel{\includegraphics{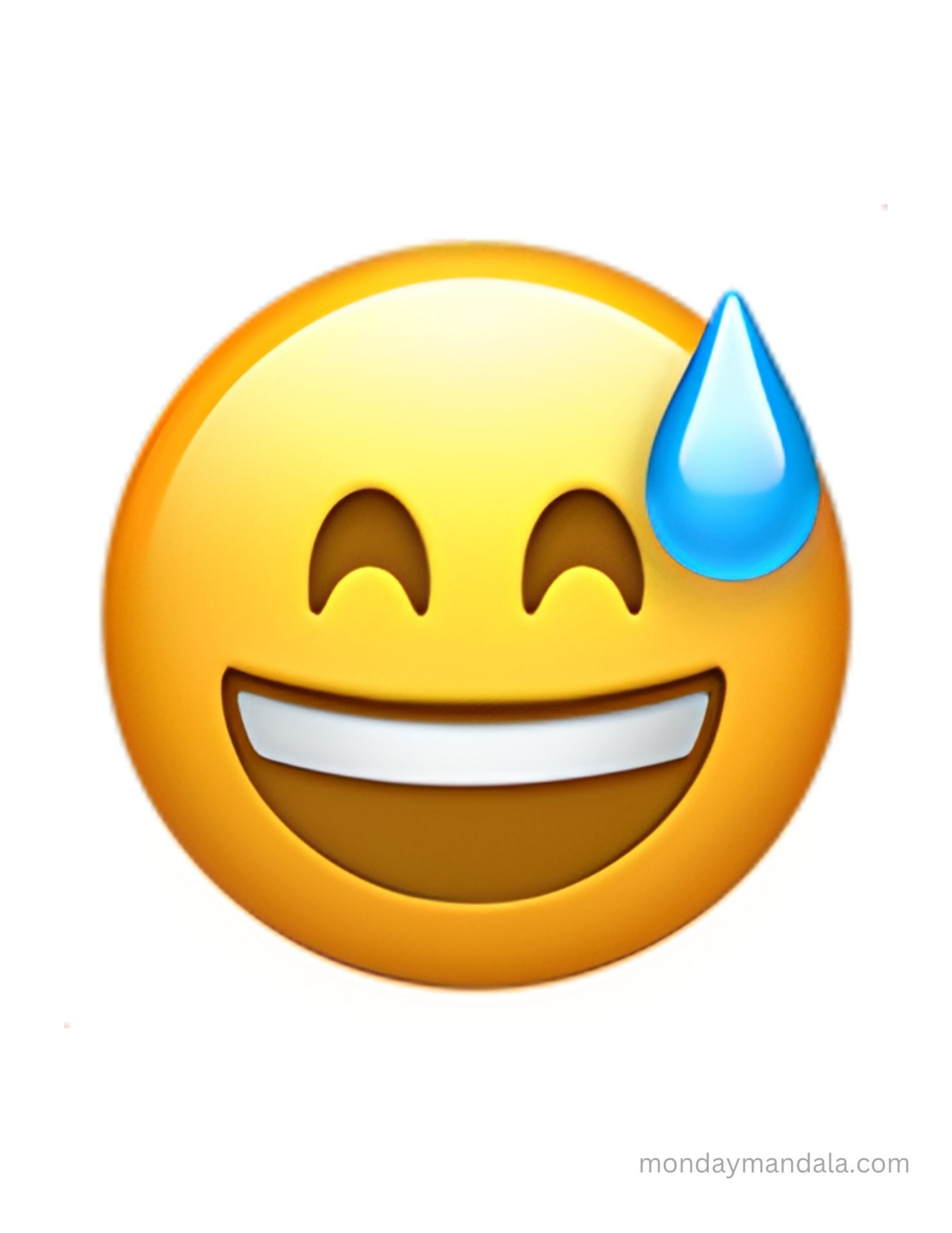}}{\strut}}
\newcommand{\emojiBeamingFaceWithSmilingEyes}{\scalerel{\includegraphics{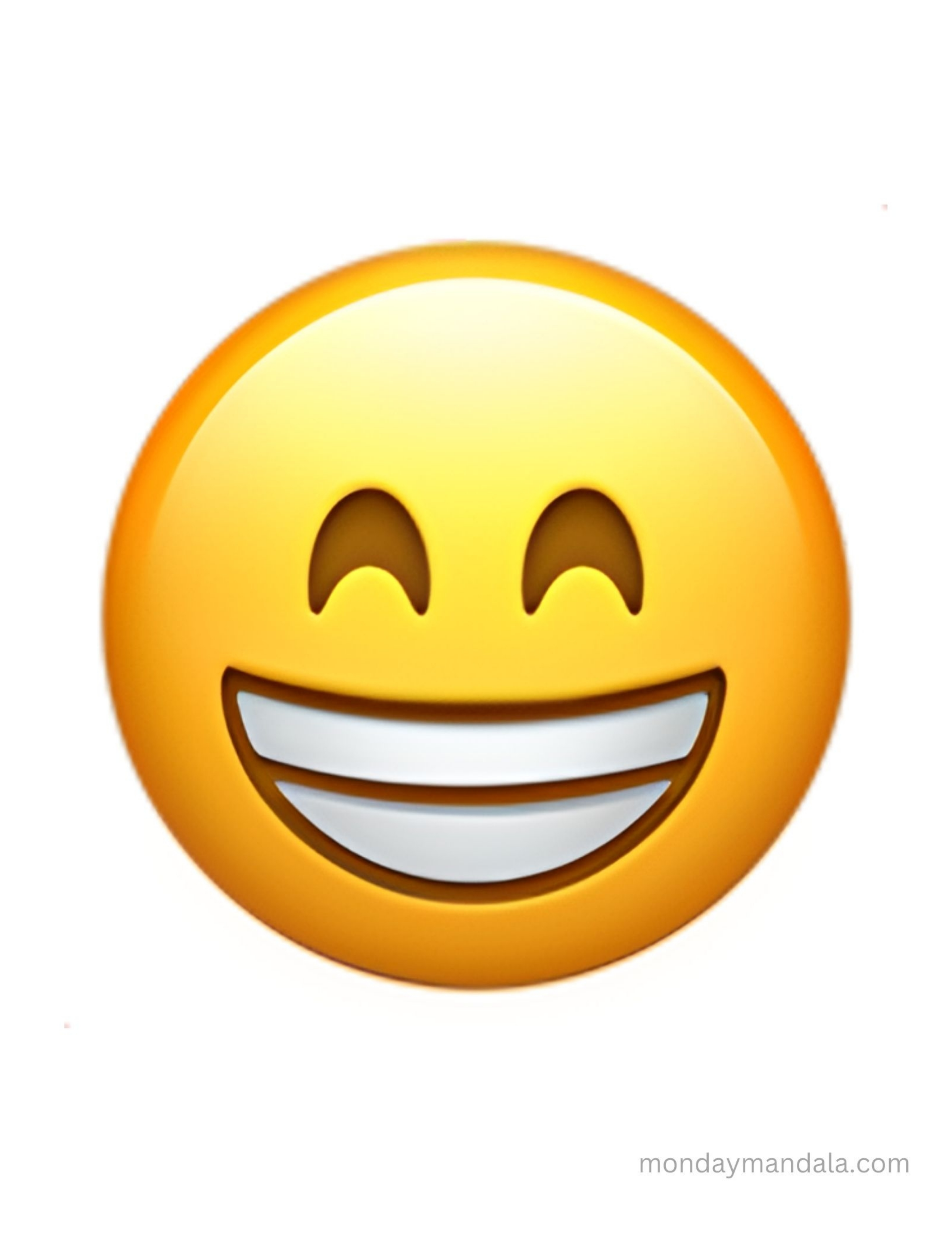}}{\strut}}
\newcommand{\emojiGrinningFace}{\scalerel{\includegraphics{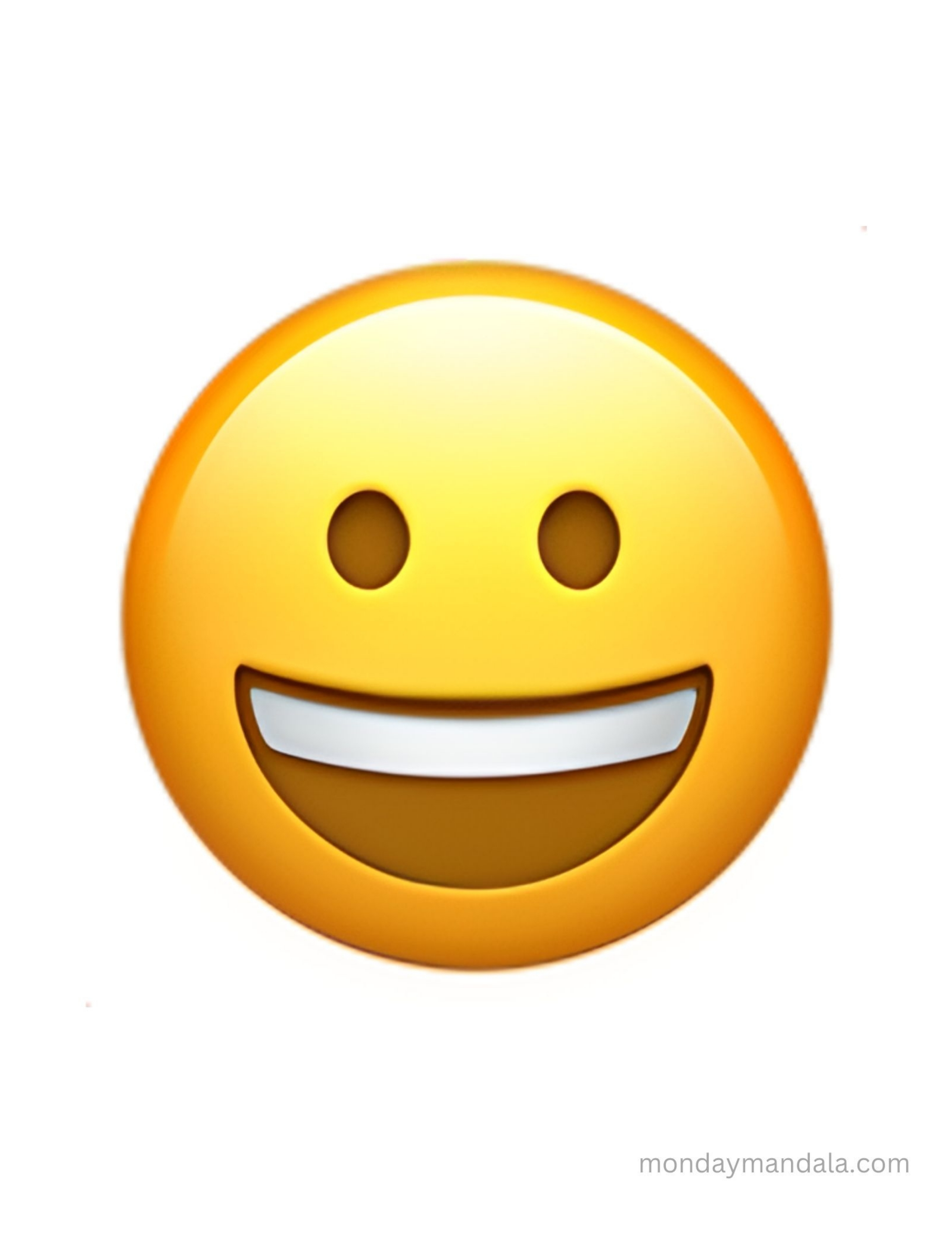}}{\strut}}
\newcommand{\emojiClappingHands}{\scalerel{\includegraphics{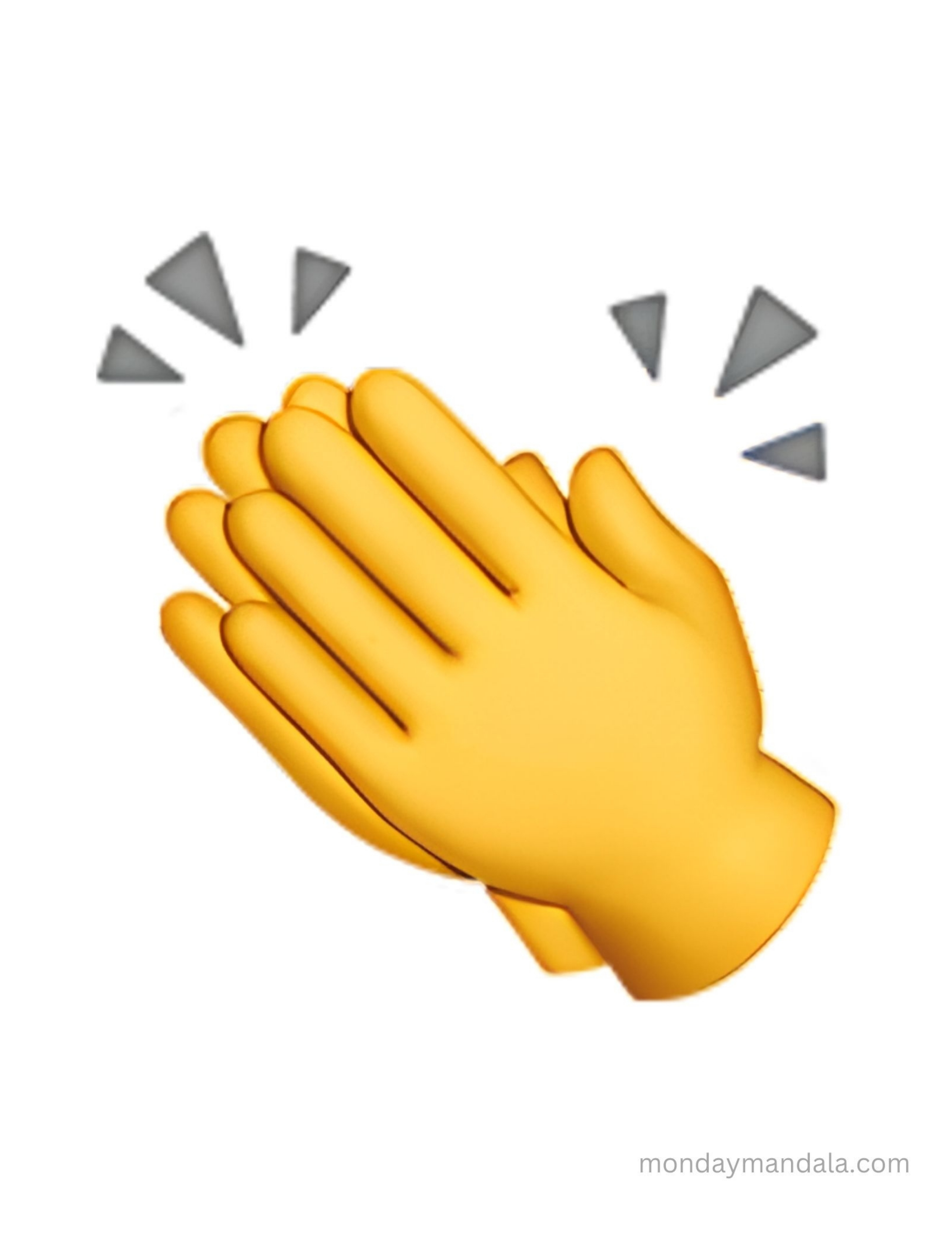}}{\strut}}

\title{Emotion Recognition for Low-Resource Turkish: Fine-Tuning BERTurk on TREMO and Testing on Xenophobic Political Discourse}

\author{
 Darmawan Wicaksono* \\
  New Media and Communications\\
  Social Sciences University of Ankara (ASBU)\\
  Ankara, Türkiye \\
  \texttt{darmawan.wicaksono@student.asbu.edu.tr} \\
   \And
 Hasri Akbar Awal Rozaq \\
  Department of Computer Science\\
  Gazi University\\
  Ankara, Türkiye \\
  \texttt{hakbar.rozaq@gazi.edu.tr} \\
  \And
 Nevfel Boz \\
  New Media and Communications\\
  Social Sciences University of Ankara (ASBU)\\
  Ankara, Türkiye \\
  \texttt{nevfel.boz@asbu.edu.tr} \\
}

\begin{document}
\maketitle
\begin{abstract}
Social media platforms like X (formerly Twitter) play a crucial role in shaping public discourse and societal norms. This study examines the term Sessiz Istila (Silent Invasion) on Turkish social media, highlighting the rise of anti-refugee sentiment amidst the Syrian refugee influx. Using BERTurk and the TREMO dataset, we developed an advanced Emotion Recognition Model (ERM) tailored for Turkish, achieving 92.62\% accuracy in categorizing emotions such as happiness, fear, anger, sadness, disgust, and surprise. By applying this model to large-scale X data, the study uncovers emotional nuances in Turkish discourse, contributing to computational social science by advancing sentiment analysis in underrepresented languages and enhancing our understanding of global digital discourse and the unique linguistic challenges of Turkish. The findings underscore the transformative potential of localized NLP tools, with our ERM model offering practical applications for real-time sentiment analysis in Turkish-language contexts. By addressing critical areas including marketing, public relations, and crisis management, these models facilitate improved decision-making through timely and accurate sentiment tracking. This highlights the significance of advancing research that accounts for regional and linguistic nuances. 
\end{abstract}


\section{Introduction}
In the digital era, social media platforms have burgeoned as arenas for public discourse, significantly influencing both societal norms and individual behaviors (Han et al., 2021). X (formerly known as Twitter, hereafter referred to as X), in particular, stands out as a vital space for real-time communication, where users from diverse backgrounds converge to share opinions and disseminate information (Kavitha et al., 2019). The term ‘sessiz istila’, translated as ‘silent invasion’ gained prominence in Turkish social media following the release of Hande Karacasu’s short film of the same name, ‘Sessiz Istila’ or ‘Silent Invasion’ \cite{bahadır2023}. ‘Silent Invasion’ is a controversial term used by critics of immigration policies to describe the perceived impact of refugees on a culture, economy, or social structure, with anti-immigration rhetoric. In Türkiye, the term highlights the socio-political and cultural shifts brought about by the influx of refugees especially Syrian refugees, reflecting the broader anti-refugee sentiment within the nation \cite{erbaysal2023}. A nuanced understanding of the emotional underpinnings of such potent terms on social media platforms, including X, can provide invaluable insights into prevailing public sentiment \cite{tan2014}, critical for policymakers, scholars, and the general public navigating these complex challenges.

This is particularly relevant for languages less represented in the digital domain, such as Turkish \cite{Toçoglu2019}, where nuances can be difficult to capture. Transfer Learning, specifically Bidirectional Encoder Representations from Transformers (BERT) models, offers a promising solution to the challenges faced by traditional methods of analyzing emotions in language-specific contexts \cite{masarifoglu2021}. As demonstrated by Nozza et al. (2020)\cite{nozza2020}, BERT models provide a confident approach to overcoming these challenges. BERT's capacity to comprehend the context of words in search queries significantly enhances the performance of natural language processing (NLP) tasks, such as emotion analysis in particular languages \cite{acheampong2021}. BERTurk, a specialized pre-trained computational model designed for the Turkish language \cite{schweter2020}, has undergone pre-training on Turkish corpora  \cite{yildirim2024, yucalar2023}, enhancing its proficiency in comprehending and processing the language while recognizing its distinctive linguistic attributes and subtleties. The optimization of BERTurk for the Turkish language makes it significantly more effective for tasks involving Turkish text compared to the general-purpose BERT model  \cite{budur2020, türkmen2023}.

The aim of this research is to analyze emotions in Turkish text by building and training various machine learning models, including deep learning networks. Furthermore, we will utilize BERTurk for NLP tasks involving Turkish text. This model provides out-of-the-box capabilities for understanding Turkish language content. The TREMO dataset was used to fine-tune the Turkish emotion analysis model \cite{tocogluandalpkocak2018}, categorizing Turkish emotions into six distinct categories: happiness, fear, anger, sadness, disgust, and surprise, based on Paul Ekman's model of basic emotions \cite{ekman1999}. The study focuses on X conversations that took place between June 2021 and December 2022, discussing the ‘Sessiz Istila’ narrative.

According to Aka Uymaz \& Kumova Metin (2023)\cite{Aka2023}, culture and language are closely intertwined in expressing emotions. Individuals from the same culture who speak the same language tend to exhibit emotions in comparable ways. However, due to structural disparities in languages and cultural differences in language usage, emotion detection models or methods that are effective for one language may not be as effective for other languages. Turkish is renowned for its agglutinative nature, distinguishing it from other languages in the field. In Turkish, a root word can convey both positive and negative meanings by changing suffixes.

This study aims to fill the gaps by (1) developing a strong, context-aware model specific to the Turkish language, utilizing BERTurk and TREMO to establish advanced Emotion Recognition Models (ERMs) for Turkish. These ERMs represent the state of the art in emotion recognition within the Turkish language, introducing a novel approach to capturing nuanced emotional expressions. (2) The model will be applied to a large-scale X data analysis related to sessiz istila. The objective is to reveal the emotional aspects of public discourse surrounding this topic, providing valuable insights into the collective sentiment of Turkish-speaking X users. Our work demonstrates the application of advanced NLP techniques to understand societal trends and sentiments in languages beyond English, significantly contributing to the field of computational social science. This enriches our understanding of global digital discourse and addresses the unique challenges presented by the agglutinative nature of Turkish.

The rest of the paper is organized as follows: Section 2 reviews previous studies on Sessiz Istila and Turkish-language emotion analysis. Section 3 outlines the methods used to analyze emotions in the Sessiz Istila corpus. Section 4 details the experimental setup and results. Section 5 discusses the findings of emotion analysis. Section 6  presents the conclusion.

\section{Material}
The short film Sessiz Istila, released on YouTube on May 3, 2022, has garnered over 3 million views. Bahadır Türk (2023)\cite{bahadır2023} utilized Critical Discourse Analysis (CDA) to examine the data from Turkish political party manifestos in relation to these short films, uncovering a blend of anxiety, fear, and anger that underscores the public’s dissatisfaction with refugees. Complementing this research, Erbaysal Filibeli \& Öneren Özbek (2023)\cite{erbaysal2023} analyzed YouTube comments on Sessis Istila through content analysis with the aid of NVivo Qualitative Research Analysis Software, identifying a dominant call for banning refugees to Türkiye and urging for action or violence against them. Their findings show that 90\% of the comments advocate for the deportation of refugees.

Building on these insights, prior studies have employed discourse analysis, extending it to emotion. Kapucu et al. (2021)\cite{kapucu2021} developed normative ratings for a large set of Turkish words across two major emotional dimensions -arousal and valence- and five basic emotions: happiness, sadness, anger, fear, and disgust. This extensive study involved translating Affective Norms for English Words into Turkish and expanding the dataset with Turkish Word Norms, resulting in 2,031 Turkish words rated by 1,527 participants. The research highlights the correlations among emotions, particularly noting the strong association between valence and happiness, as well as arousal and fear.

Advancements in the field have introduced modern methodologies leveraging machine learning for emotion analysis. Boynukalin \& Karagoz (2013)\cite{boynukalin2013} demonstrated the efficacy of the Complementary Naive Bayes technique and the ISEAR dataset, achieving an 81.34\% accuracy rate in identifying four basic emotions. Similarly, Tocoglu and Alpkocak (2014, 2018)\cite{tocoglu2014, tocogluandalpkocak2018} explored emotional state analysis and sentiment analysis, culminating in the creation of the TREMO dataset, which significantly enhances the precision of machine learning algorithms, including support vector machines (SVMs), in emotion analysis. Further building on their foundation, \cite{Toçoglu2019} developed a Turkish emotion lexicon based on the TREMO dataset. Showed that the use of the proposed lexicon for emotion analysis produced a comparable result with machine learning algorithms.

Further exploring machine learning, Doğan \& Kaya (2019)\cite{doğan2019} compared the effectiveness of Word2Vec and FastText for sentiment analysis in Turkish, reporting significant improvements, with FastText showing superior performance. Deep learning techniques, notably the BERT \cite{devlin2019} models and its Turkish variant, BERTurk \cite{schweter2020}, have also shown promise in processing the Turkish language, enhancing Natural Language Processing (NLP) tasks’ performance. Koksal (2021)\cite{koksal2021} and subsequent studies by Aka Uymaz and Kumova Metin (2023)\cite{Aka2023}; Yildirim (2024)\cite{yildirim2024}; Yucalar (2023)\cite{yucalar2023} have further validated the superior capabilities of BERT-based models in NLP, emotion, and sentiment analysis, especially in a language as complex as Turkish. In contrast to static word embeddings, which assign a uniform vector to each word irrespective of context, BERT score employs the BERT model to assess the semantic alignment between reference and generated texts. The score is calculated by computing the cosine similarity between two sets of embeddings. A higher cosine similarity indicates a higher BERT score and greater text similarity \cite{zhang2019}.

Recent evaluations, including those by Onan \& Balbal (2024)\cite{Onan2024} and Dehghan \& Yanikoglu (2024)\cite{dehghan2024}, highlight the superior adaptability of BERT and its variants in NLP tasks like hate speech detection, often rivaling ChatGPT in performance. Similarly, Altmay \& Turhan (2022)\cite{altmay2022} emphasize the growing sophistication of methodologies in Turkish emotion analysis, with approaches spanning machine learning, deep learning, and lexicon-based techniques. This body of research not only enhances understanding of human emotions in digital communication but also advances sentiment analysis in underrepresented languages like Turkish, addressing socio-political discourse and expanding NLP’s methodological horizons.

\section{Methods}
\label{sec:headings}
This section presents an overview of the processes employed in the empirical analysis, including data collection, data cleaning and normalization, data splitting, fine-tuning, and emotion analysis, as can be seen in Figure 1.

\begin{figure}[htbp]
  \centering
  \includegraphics[width=0.8\linewidth]{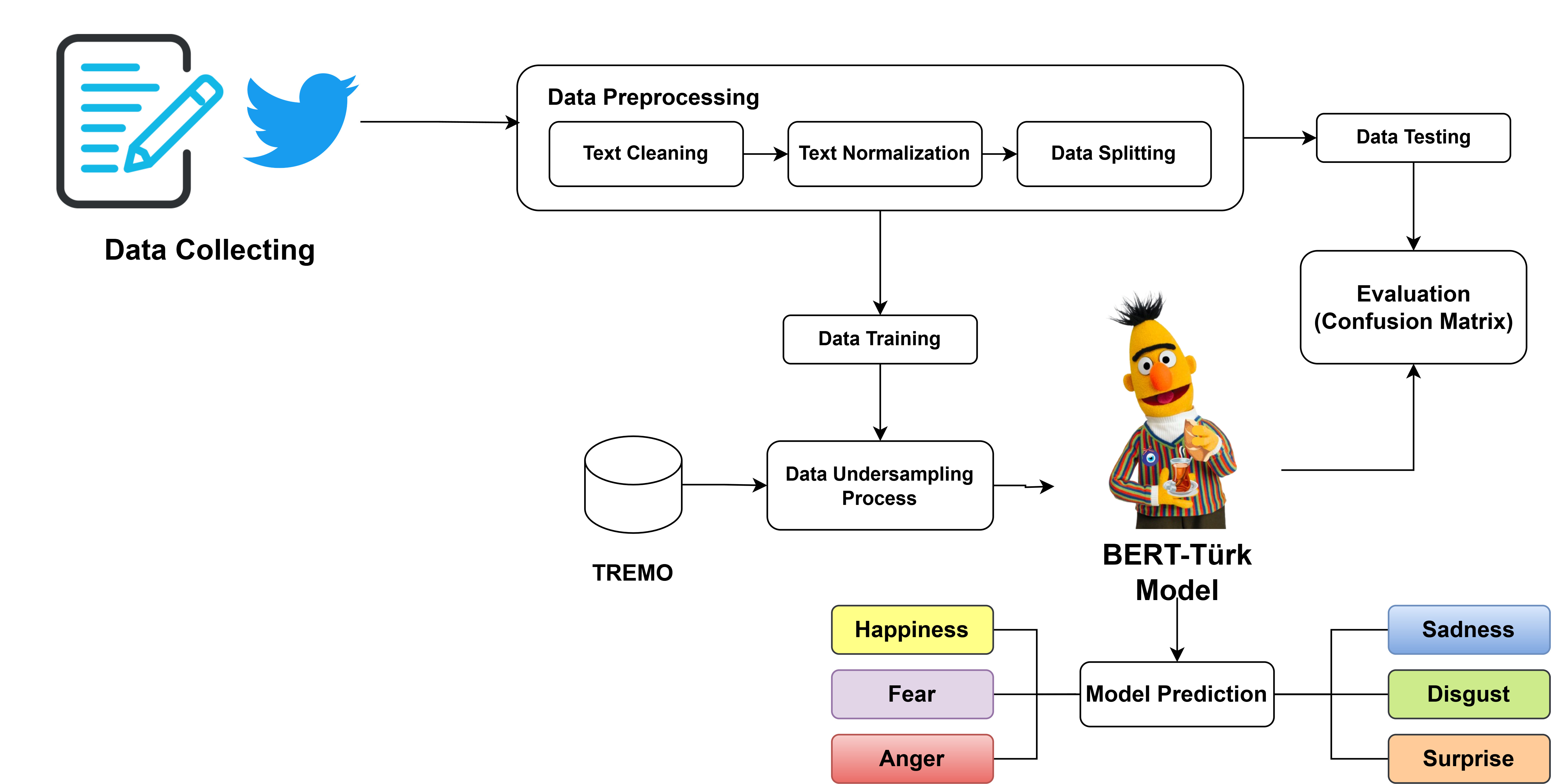} 
  \vspace{1em} 
  \caption{Research flow.}
  \label{fig:fig1}
\end{figure}

\subsection{Data collection}
This study utilized a comprehensive dataset of X (formerly known as Twitter, hereafter referred to as X) conversations spanning from June 1, 2021 to December 31, 2022. The conversations were specifically identified through the keyword ‘sessiz istila’. Data was collected using academictwitteR API \cite{barrie2021}, data taken on January 26, 2022, when free Twitter API was still available \cite{barnes2023}, adhering to the platform’s data usage policies and ethical guidelines concerning user privacy and data anonymization. Tweets were filtered to include only those publicly available and written in Turkish, resulting in a dataset comprising 47,024 tweets.

\subsection{Data cleaning and normalization}
Pota et al. (2021)\cite{pota2021} proposed a method for normalizing X tweets downloaded from the Twitter API. The proposed approach replaces retweets, URLs, user mentions, hashtags, and emojis with normalized forms, while retaining punctuation to aid in identifying the tweet’s structure, as shown in Table 1. The normalization using following rules based on R programming language:

\begin{verbatim}
"RT @\\w+:", "(retweetlemek)" --> retweets  
"\\w+://\\S+", "(bağlantı adresi)" --> URLs  
"@[a-z,_,A-Z,0-9]*", "(kullanıcı)" --> user mentions  
"#(\\w+)", "(\\1)" --> hashtags
\end{verbatim}

\begin{table}[htbp]
 \caption{Tweet normalization replaces retweets, URLs, user mentions, hashtags, and emoticons (\protect\emojiClappingHands{}, \protect\emojiGrinningFaceWithBigEyes{}, \protect\emojiGrinningFaceWithSmilingEyes{}, \protect\emojiGrinningSquintingFace{}, \protect\emojiGrinningFaceWithSweat{} and others) with their normal forms.}
 \label{tab:normalization}  
  \centering
  \begin{tabular}{p{0.5cm} p{14cm}}
    \addlinespace
    \toprule
    No     & Original \\
    \midrule
    1 & RT @Zalim\_Fira: Sessiz istila \texttt{https://t.co/tB5KS6U94u} \texttt{https://t.co/2mg9KzOQhu} \\ 
    \addlinespace
    2 & RT @SedefKabas: Sessiz istila… Sarsıcı gerçek… Sinsi işgal… \#sessizistila \#suriyeliler \#sığınmacısorunu \#afganlar \#araplar \#afrikalılar \#… \\
    \addlinespace
    3 & @KaracasuHande Çok yazık. Sessiz istila 5 gelirse sonunda RTE olursa hiç şaşırmayacağım. \\
    \addlinespace
    4 & @KaracasuHande \emojiClappingHands{} \emojiClappingHands{} \emojiClappingHands{} \emojiClappingHands{} \emojiClappingHands{}baş tacısın sessiz istila \\
    \addlinespace
    \toprule
    No     & Normalized \\
    \midrule
    1 & (retweetlemek) sessiz istila (bağlantı adresi) (bağlantı adresi) \\ 
    \addlinespace
    2 & (retweetlemek) sessiz istila… sarsıcı gerçek… sinsi işgal… (sessizistila) (suriyeliler) (sığınmacısorunu) (afganlar) (araplar) (afrikalılar) … \\
    \addlinespace
    3 & (kullanıcı) çok yazık. sessiz istila 5 gelirse sonunda rte olursa hiç şaşırmayacağım. \\
    \addlinespace
    4 & (kullanıcı) (elleri çırpmak)(elleri çırpmak)(elleri çırpmak)(elleri çırpmak)(elleri çırpmak)baş tacısın sessiz istila \\
    \bottomrule
  \end{tabular}
\end{table}

Normalizing emojis in Turkish presents certain challenges, as researchers have not found any method to translate emoji forms into Turkish with the same degree of completeness as in English. Consequently, the researcher has developed its own emoji dataset \cite{wicaksono2024}, initially based on emojis standards \cite{emojiterra2024}, which was subsequently translated into Turkish, as shown in Table 2.

\begin{table}[htbp]
 \caption{Turkish-emoji translation.}
 \label{tab:turkish-emoji}  
  \centering
  \begin{tabular}{p{3.2cm} p{1cm} p{5cm} p{4.5cm}}
    \addlinespace
    \toprule
    Unicode.Code.Point.s.     & Emoji      & Description      & Translated   \\
    \midrule
    U+1F600 & \emojiGrinningFace{} & Grinning Face & Sırıtan Yüz \\ 
    \addlinespace
    U+1F603 & \emojiGrinningFaceWithBigEyes{} & Grinning Face With Big Eyes & İri Gözlü Sırıtan Yüz \\
    \addlinespace
    U+1F604 & \emojiGrinningFaceWithSmilingEyes{} & Grinning Face With Smiling Eyes & Gülen Gözlerle Sırıtan Yüz \\
    \addlinespace
    U+1F601 & \emojiBeamingFaceWithSmilingEyes{}  & Beaming Face With Smiling Eyes & Gülen Gözlerle Işıldayan Yüz \\
    \addlinespace
    U+1F606 & \emojiGrinningSquintingFace{}  & Grinning Squinting Face & Sırıtan şaşı yüz \\
    \addlinespace
    U+1F605 & \emojiGrinningFaceWithSweat{}  & Grinning Face With Sweat & Terli Sırıtan Yüz \\
    \bottomrule
  \end{tabular}
\end{table}

Following this, the researcher developed a Python program that could convert all emojis from an X dataset into Turkish for further analysis. The text normalization process removes unwanted characters and converts all characters to lowercase. Duplicate tweets are intentionally kept as they are crucial for this research. Stieglitz \& Dang-Xuan (2013)\cite{stieglitz2013} have proven that emotionally charged X messages are significantly more likely to be retweeted than neutral messages. Additionally, Buckley et al. (2012)\cite{buckley2012} have found a strong positive correlation between the emotions expressed in articles and the physiological arousal they cause, ultimately resulting in increased sharing. Alzahrani \& Jololian (2021)\cite{alzahrani2021} demonstrated that BERT achieves the highest accuracy without pre-processing techniques. The study found that removing stop words negatively impacted the performance of the BERT model. It is important to note that pre-trained models require all tokens and larger texts to learn from; therefore, no stop words will be removed.

\subsection{Splitting data}
In machine learning model training, bias is often linked to over-training or over-fitting \cite{reitermanová2010}. While no universal rule exists for the optimal training-to-testing data split, Muraina (2022)\cite{muraina2022} highlights that a 90/10 division is frequently effective, depending on dataset characteristics. Our study corroborates this, finding that a 90/10 split yielded optimal results. Accordingly, 90\% of the dataset was used for training and 10\% for testing, ensuring the model was extensively trained while maintaining rigorous evaluation on an independent test set, thereby validating its ability to recognize and classify emotions in the Turkish context accurately.

\subsection{Fine-tune}
To fine-tune a Turkish-language model, this study utilizes the TREMO dataset, specifically designed for emotion analysis in the Turkish context. The dataset has been normalized to lowercase for consistency and balance to ensure equal distribution across six emotion categories, as outlined in Table 3, thereby minimizing bias and enhancing the reliability of the model’s performance.

\begin{table}[htbp]
 \caption{Emotion labels.}
 \label{tab:emotion-labels}  
  \centering
  \begin{tabular}{p{1.5cm} p{1.5cm}}
    \addlinespace
    \toprule
    Emotion     & Label \\
    \midrule
    Happy & 0 \\ 
    \addlinespace
    Fear & 1 \\
    \addlinespace
    Sadness & 2 \\
    \addlinespace
    Disgust & 3 \\
    \addlinespace
    Surprise & 4 \\
    \addlinespace
    Anger & 5 \\
    \bottomrule
  \end{tabular}
\end{table}

The TREMO dataset is tokenized using the ‘bert-base-turkish-cased’ model from the BERTurk tokenizer, pre-trained on Turkish data, and formatted into tensor-compatible structures for efficient training and evaluation. The model was trained for three epochs, following the training strategy proposed by Devlin et al. (2019)\cite{devlin2019}, to ensure optimal performance and observe accuracy trends across epochs. Classification accuracy was further assessed using a confusion matrix, offering a detailed breakdown of predictions essential for evaluating model performance \cite{heydarian2022, subarkah2024}. This method facilitates the development of an advanced Emotion Recognition Model (ERM) for Turkish, aiming to set a new benchmark in Turkish emotion recognition.

\subsection{Emotion analysis}
Emotion analysis was performed using a GPU on Google Colab, leveraging our novel Emotion Recognition Model (ERM) for emotion classification. The \texttt{"predict\_emotions"} function incorporated a threshold parameter to ensure reliable predictions, setting a minimum confidence level of 0.6 for emotion classification. Initially, the model generated raw logits, representing unprocessed scores for each emotion class, which were converted to probabilities using torch.softmax. The highest probability for each prediction, obtained via torch.max, was compared against the threshold. Predictions falling below this threshold were marked as ambiguous with a value of -1, ensuring only confident predictions were labeled with specific emotions. This approach enhanced classification reliability by excluding uncertain predictions and maintaining a robust confidence level in the assigned emotion categories.

\section{Results }
\label{sec:others}
The results section provides a detailed overview of the research process and its findings, adhering to ethical standards for social media research by safeguarding the privacy and anonymity of X users. All data was meticulously anonymized, ensuring no personal identifiers were included in the dataset.

\subsection{Balancing Trainer Data}
To effectively utilize the TREMO dataset for fine-tuning BERTurk, it is first necessary to balance the dataset’s distribution of emotions. As illustrated in Figure 2, the initial distribution of emotions in TREMO is uneven, with ‘surprise’ being the least represented category, comprising 3,003 sentences. To ensure uniformity, we set 3,003 sentences as the limit for each of the remaining five emotions, thus creating a balanced dataset as shown in Figure 3. Notably, only instances labeled as ValidatedEmotion within the TREMO dataset were included in this balancing process, ensuring that only confirmed emotion data is included before equalizing all emotions to this 3,003 sentence threshold.

\begin{figure}[htbp]
  \centering
  \includegraphics[width=0.8\linewidth]{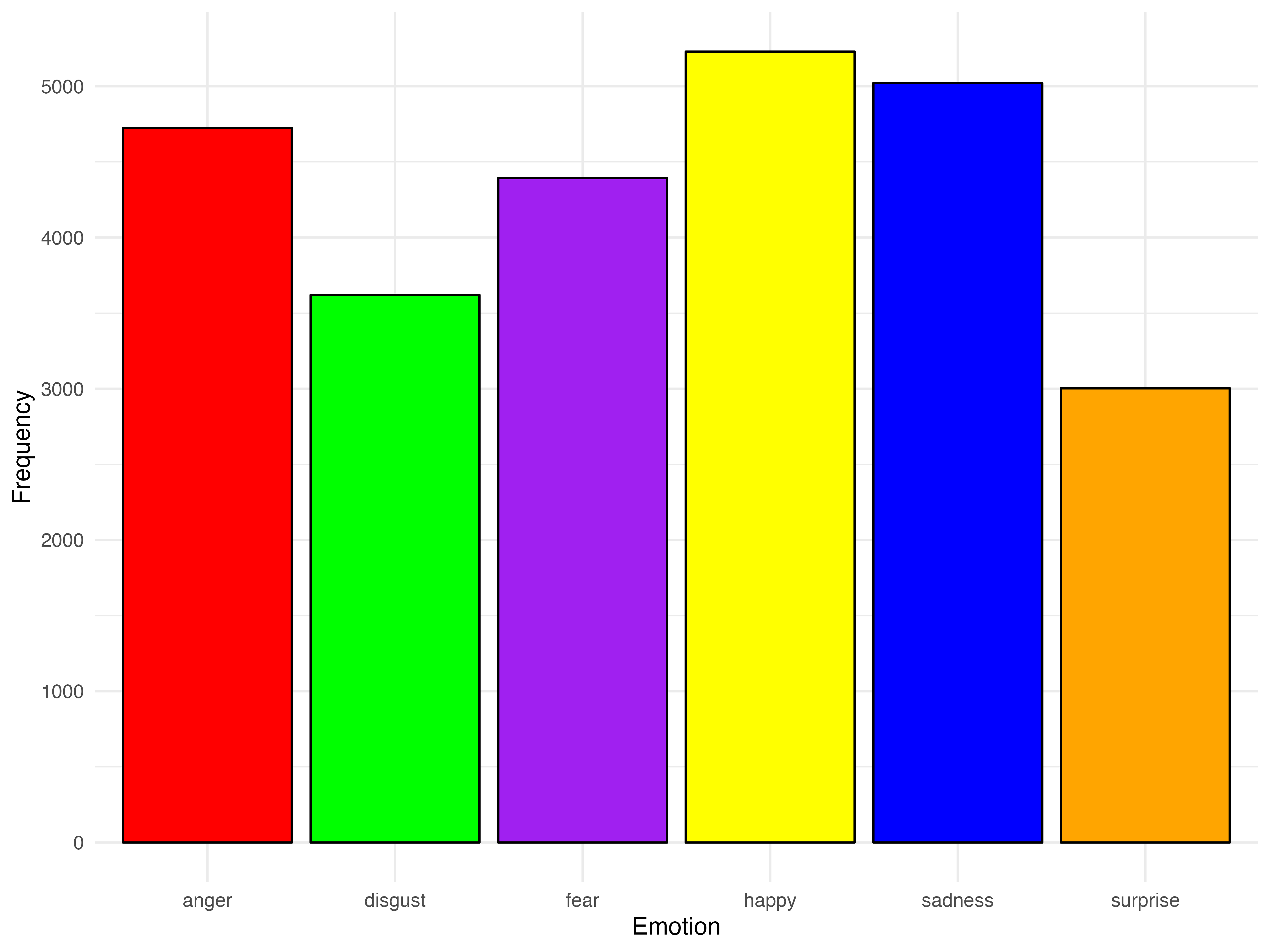} 
  \vspace{1em} 
  \caption{Initial emotion dataset in TREMO.}
  \label{fig:fig2}
\end{figure}

\begin{figure}[htbp]
  \centering
  \includegraphics[width=0.8\linewidth]{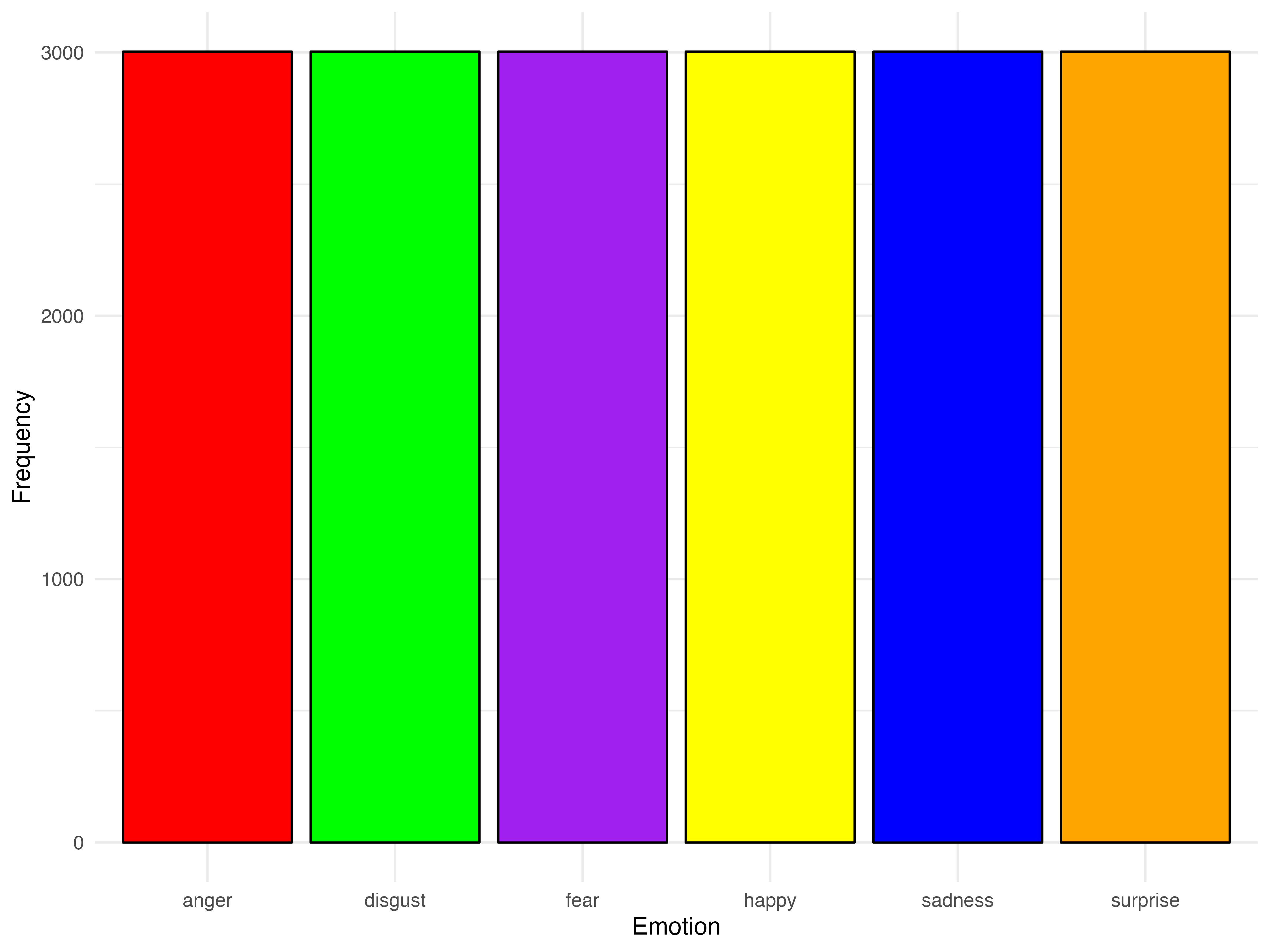} 
  \vspace{1em} 
  \caption{Calibrated dataset in TREMO.}
  \label{fig:fig3}
\end{figure}

\subsection{Model Performance}
This study highlights the efficacy of an emotion recognition model, evidenced by low training and validation losses, high validation accuracy and precision, and minimal discrepancies between phases, underscoring robust learning and generalization without overfitting. Its nuanced ability to differentiate complex emotional categories further affirms its potential for precise emotion classification in practical applications.

\begin{figure}[htbp]
  \centering
  \includegraphics[width=0.8\linewidth]{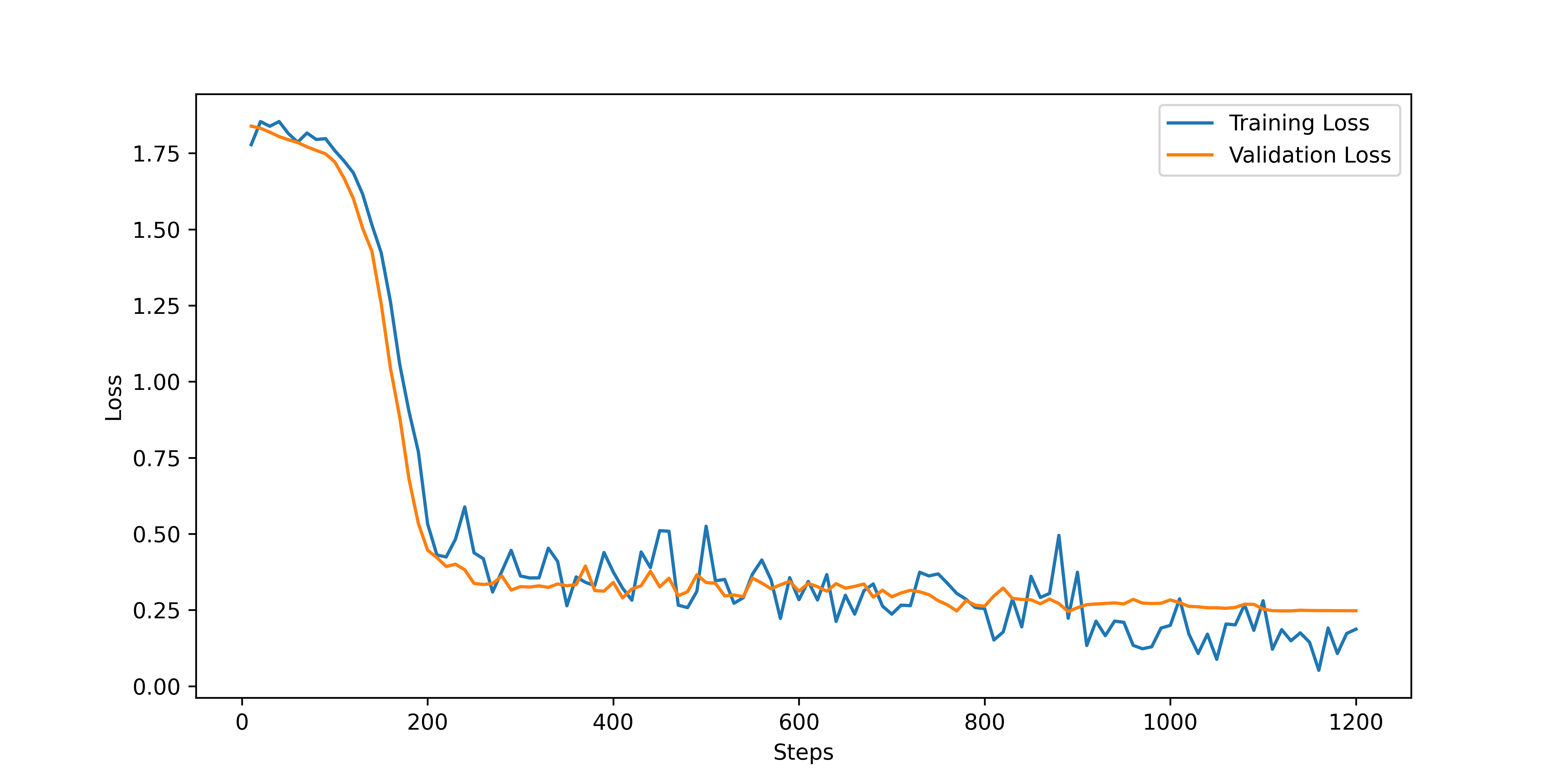} 
  \vspace{1em} 
  \caption{Training and validation loss convergence.}
  \label{fig:fig4}
\end{figure}

Figure 4 depicts the training and validation loss progression over iterations, with x-axis showing training steps and the y-axis representing loss values as a measure of prediction error. Initially, both losses decrease sharply, reflecting rapid adaptation as the model minimizes errors. By around the 200th step, losses stabilize, signalling convergence and diminishing returns from further training. The close alignment of training and validation losses at this stage highlights effective generalization with minimal overfitting. Eventually, both losses plateau at low values, indicating a well-trained model that balances robustness and generalization, achieving optimal performance with minimal overfitting.

\begin{figure}[htbp]
  \centering
  \includegraphics[width=0.8\linewidth]{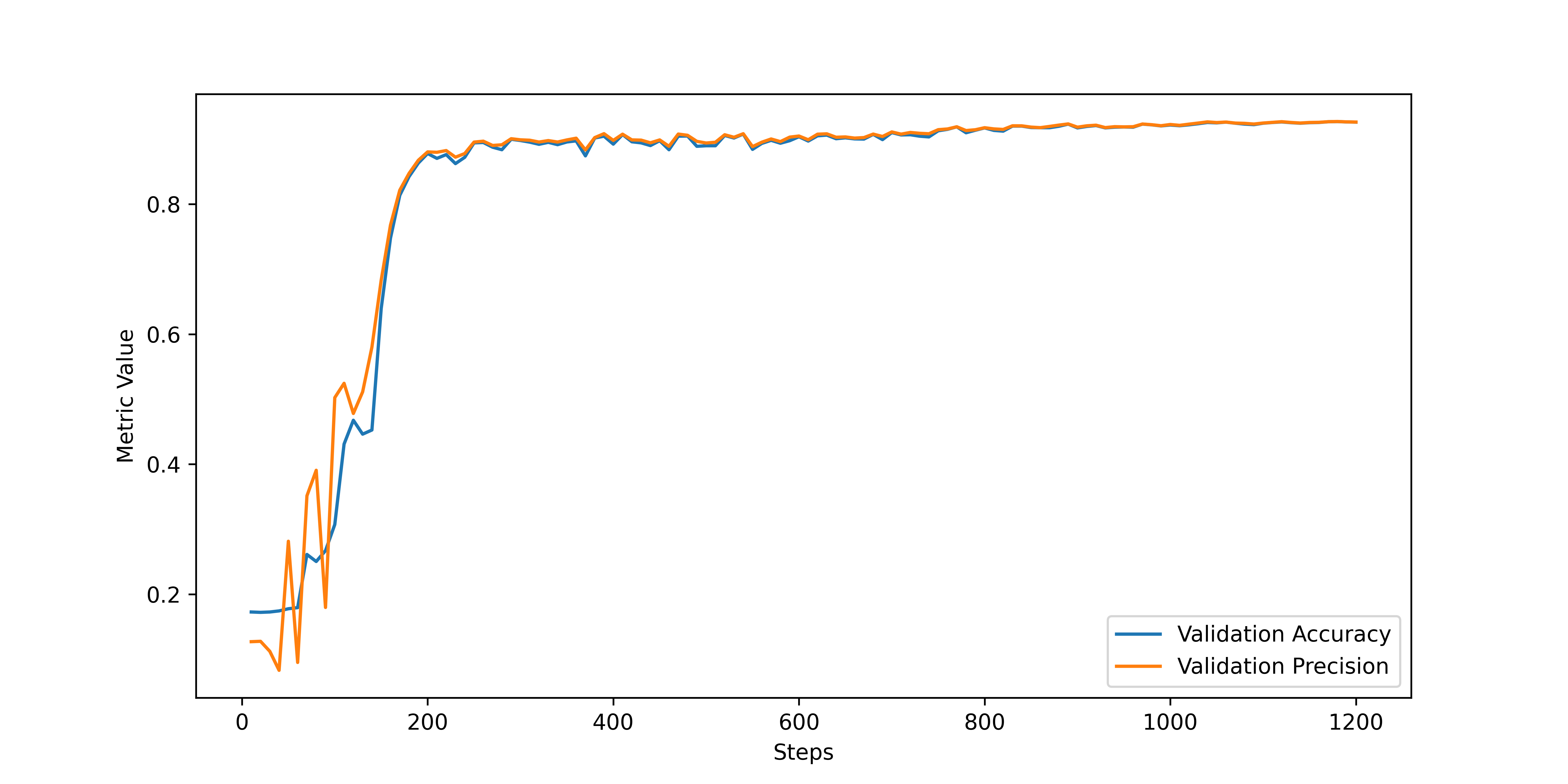} 
  \vspace{1em} 
  \caption{Model convergence in validation accuracy and precision.}
  \label{fig:fig5}
\end{figure}

Figure 5 illustrates the progression of validation accuracy and precision during training, showcasing the model’s ability to classify accurately while minimizing false positives. Both metrics rise rapidly in the initial stages, reflecting effective learning, and stabilize around 200 training steps, indicating convergence and optimal performance on the validation set. The alignment between accuracy and precision underscores the model’s balanced capability to correctly classify emotions in the Turkish-language dataset while minimizing errors. This stabilization at high values demonstrates reliable generalization, suggesting minimal gains from further training.

\begin{figure}[htbp]
  \centering
  \includegraphics[width=0.8\linewidth]{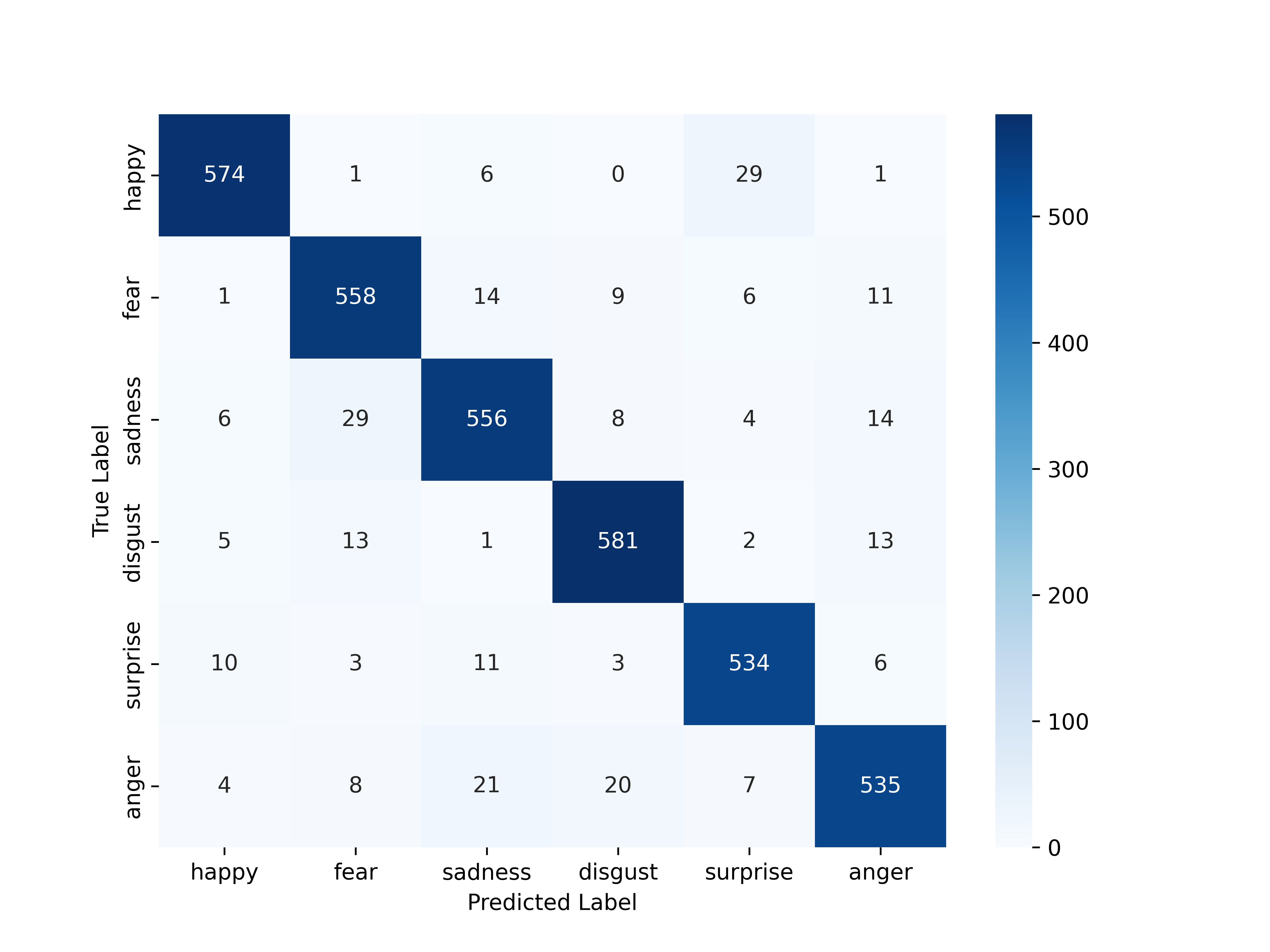} 
  \vspace{1em} 
  \caption{Confusion matrix.}
  \label{fig:fig6}
\end{figure}

The confusion matrix results, illustrated in Figure 6, demonstrate that the model achieved a high overall accuracy of 92.62\%, indicating robust performance across six emotion categories: happiness, fear, sadness, disgust, surprise, and anger. The calculation of this accuracy metric is presented below:

\begin{equation}
\textit{Accuracy} = \frac{\text{Total Correct Predictions}}{\text{Total Predictions}}
\end{equation}

The total correct predictions (diagonal sum) were 574, 558, 556, 581, 534, and 535, totalling 3338. With a total of 3605 predictions across the matrix, the accuracy of the model is:

\begin{equation}
\textit{Accuracy} = \frac{3338}{3605} \approx 0.9262 \text{ or } 92.62\%
\end{equation}

The model’s performance metrics demonstrate its reliable capacity for classification across six specified emotion categories, as evidenced by the precision, recall, and F1 scores. These metrics reflect a consistently high level of performance across categories, with precision values ranging from 0.9034 to 0.9518, recall values from 0.9115 to 0.9493, and F1 scores from 0.9091 to 0.9505.

Notably, the highest F1 scores were recorded for happy (0.9505) and disgust (0.9397) categories. Conversely, the model exhibited slightly lower performance in distinguishing sadness and anger, as indicated by an F1 score of 0.9091 and 0.9150. These findings suggest that the model is highly effective in recognizing emotions with minimal misclassification, particularly in identifying happiness and disgust.

\subsection{Model Prediction}
Building on these results, this analysis extends the findings by examining the model’s predictions on social media data from June 1, 2021, to December 31, 2022, focusing on significant emotional trends. Visualizations, such as graphs, tables, and emotional maps, highlight anger and surprise as the most dominant emotions. Monthly and yearly data further reveal shifts in public sentiment during this period, as shown in Figure 7.

\begin{figure}[htbp]
  \centering
  \includegraphics[width=0.8\linewidth]{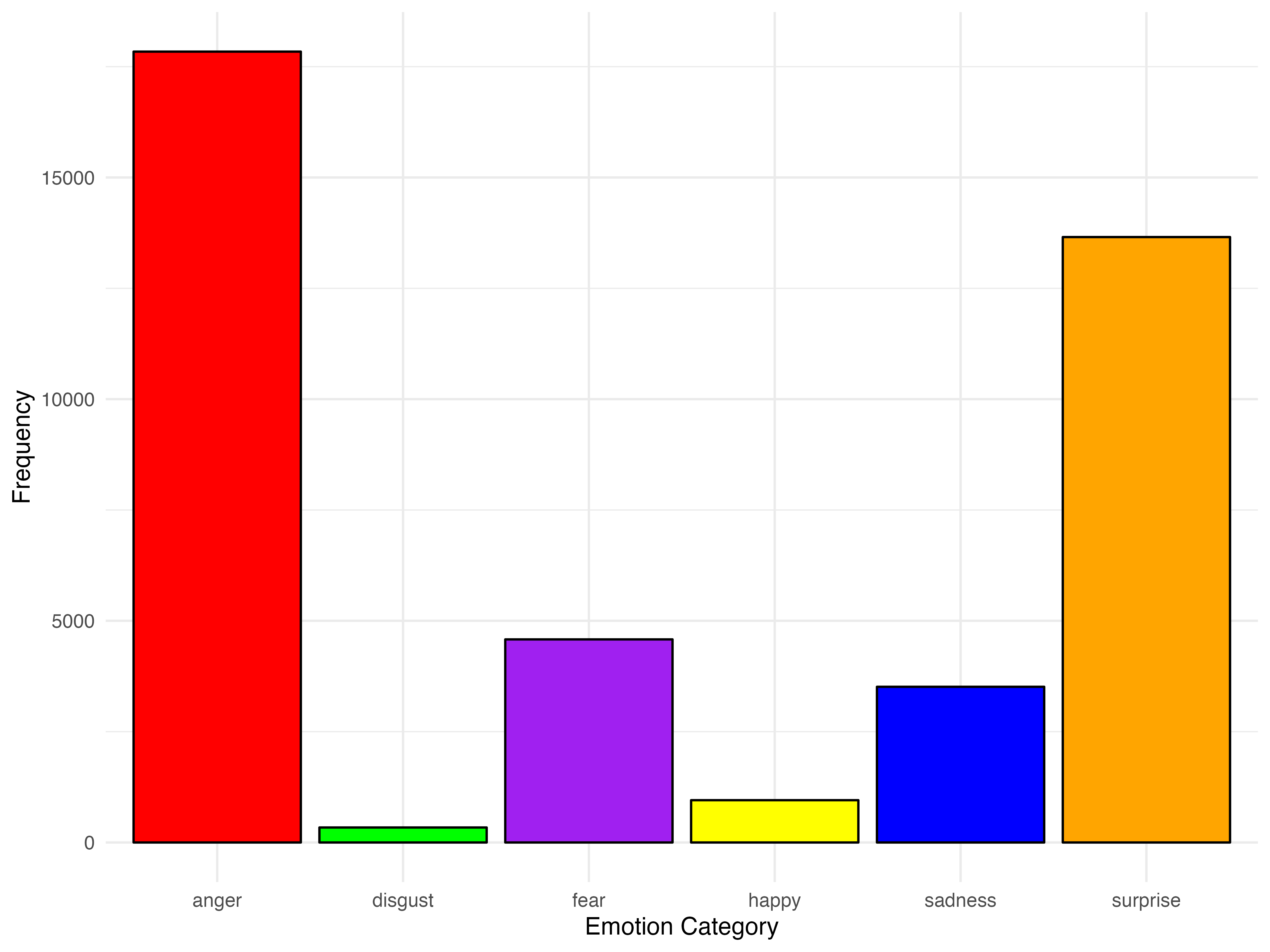} 
  \vspace{1em} 
  \caption{Emotions graph throughout 2021 to 2022.}
  \label{fig:fig7}
\end{figure}

\begin{table}[htbp]
 \caption{Percentage of emotions throughout 2021 to 2022.}
 \label{tab:2021to2022}  
  \centering
  \begin{tabular}{p{1.7cm} p{2.5cm} p{1.5cm}}
    \addlinespace
    \toprule
    Emotion     & Emotion Count      & Percentage   \\
    \midrule
    Happy & 954 & 2.3 \\ 
    \addlinespace
    Fear & 4581 & 11.2 \\
    \addlinespace
    Sadness & 3512 & 8.6 \\
    \addlinespace
    Disgust & 338 & 0.8 \\
    \addlinespace
    Surprise & 13656 & 33.4 \\
    \addlinespace
    Anger & 17839 & 43.6 \\
    \bottomrule
  \end{tabular}
\end{table}

\begin{table}[htbp]
 \caption{Percentage of emotions in 2021.}
 \label{tab:2021}  
  \centering
  \begin{tabular}{p{1.7cm} p{3cm} p{1.5cm}}
    \addlinespace
    \toprule
    Month     & Emotion Count      & Percentage   \\
    \midrule
    Happy & 12 & 0.8 \\ 
    \addlinespace
    Fear & 135 & 9.3 \\
    \addlinespace
    Sadness & 185 & 12.7 \\
    \addlinespace
    Disgust & 1 & 0.1 \\
    \addlinespace
    Surprise & 815 & 56.2 \\
    \addlinespace
    Anger & 303 & 20.9 \\
    \bottomrule
  \end{tabular}
\end{table}

\begin{table}[htbp]
 \caption{Percentage of emotions in 2022.}
 \label{tab:2022}  
  \centering
  \begin{tabular}{p{1.7cm} p{2.5cm} p{1.5cm}}
    \addlinespace
    \toprule
    Emotion     & Emotion Count      & Percentage   \\
    \midrule
    Happy & 942 & 2.4 \\ 
    \addlinespace
    Fear & 4446 & 11.3 \\
    \addlinespace
    Sadness & 3327 & 8.4 \\
    \addlinespace
    Disgust & 337 & 0.9 \\
    \addlinespace
    Surprise & 12841 & 32.6 \\
    \addlinespace
    Anger & 17536 & 44.5 \\
    \bottomrule
  \end{tabular}
\end{table}

\begin{table}[htbp]
 \caption{Percentage of peak emotions by month throughout 2021 to 2022.}
 \label{tab:month-2021to2022}  
  \centering
  \begin{tabular}{p{1.5cm} p{3cm} p{1.5cm}}
    \addlinespace
    \toprule
    Month     & Total Emotion Count      & Percentage   \\
    \midrule
    2021-06 & 6 & 0 \\ 
    \addlinespace
    2021-07 & 1203 & 2.9 \\
    \addlinespace
    2021-08 & 58 & 0.1 \\
    \addlinespace
    2021-09 & 3 & 0 \\
    \addlinespace
    2021-10 & 4 & 0 \\
    \addlinespace
    2021-11 & 3 & 0 \\
    \addlinespace
    2021-12 & 174 & 0.4 \\
    \addlinespace
    2022-01 & 12 & 0 \\ 
    \addlinespace
    2022-02 & 89 & 0.2 \\
    \addlinespace
    2022-03 & 138 & 0.3 \\
    \addlinespace
    2022-04 & 256 & 0.6 \\
    \addlinespace
    2022-05 & 21228 & 51.9 \\
    \addlinespace
    2022-06 & 8955 & 21.9 \\ 
    \addlinespace
    2022-07 & 2578 & 6.3 \\
    \addlinespace
    2022-08 & 2097 & 5.1 \\
    \addlinespace
    2022-09 & 1957 & 4.8 \\
    \addlinespace
    2022-10 & 216 & 0.5 \\
    \addlinespace
    2022-11 & 833 & 2 \\
    \addlinespace
    2022-12 & 1070 & 2.6 \\
    \bottomrule
  \end{tabular}
\end{table}

In particular, there were 17,839 tweets expressing anger. Table 4 illustrates that surprise was the second most frequent emotion, with 13,656 mentions. Overall, anger accounted for 43.6\% and surprise for 33.4\% of the total conversations between 2021 and 2022. A notable distinction is that surprise dominated in 2021, as shown in Table 5, whereas anger became more prominent in 2022, as shown in Table 6. Analysis of data from 2021 and 2022 reveals that the majority of conversations occurred in May and June of 2022, with the highest peak observed in May at 51.9\% and a subsequent peak in June at 21.9\%, as shown in Table 7. Other emotional categories, including fear, sadness, disgust, and happiness, were comparatively less prominent. Among these categories, fear and sadness were the most notable, displaying higher frequencies relative to the other categories. Specifically, fear was recorded 5,581 times, while sadness appeared 3,512 times, with both emotions being comparable in frequency. In contrast, disgust and happiness were less frequent, with only 338 and 954 occurrences, respectively, positioning these two as the least significant among the categories. 

\begin{figure}[htbp]
  \centering
  \includegraphics[width=0.8\linewidth]{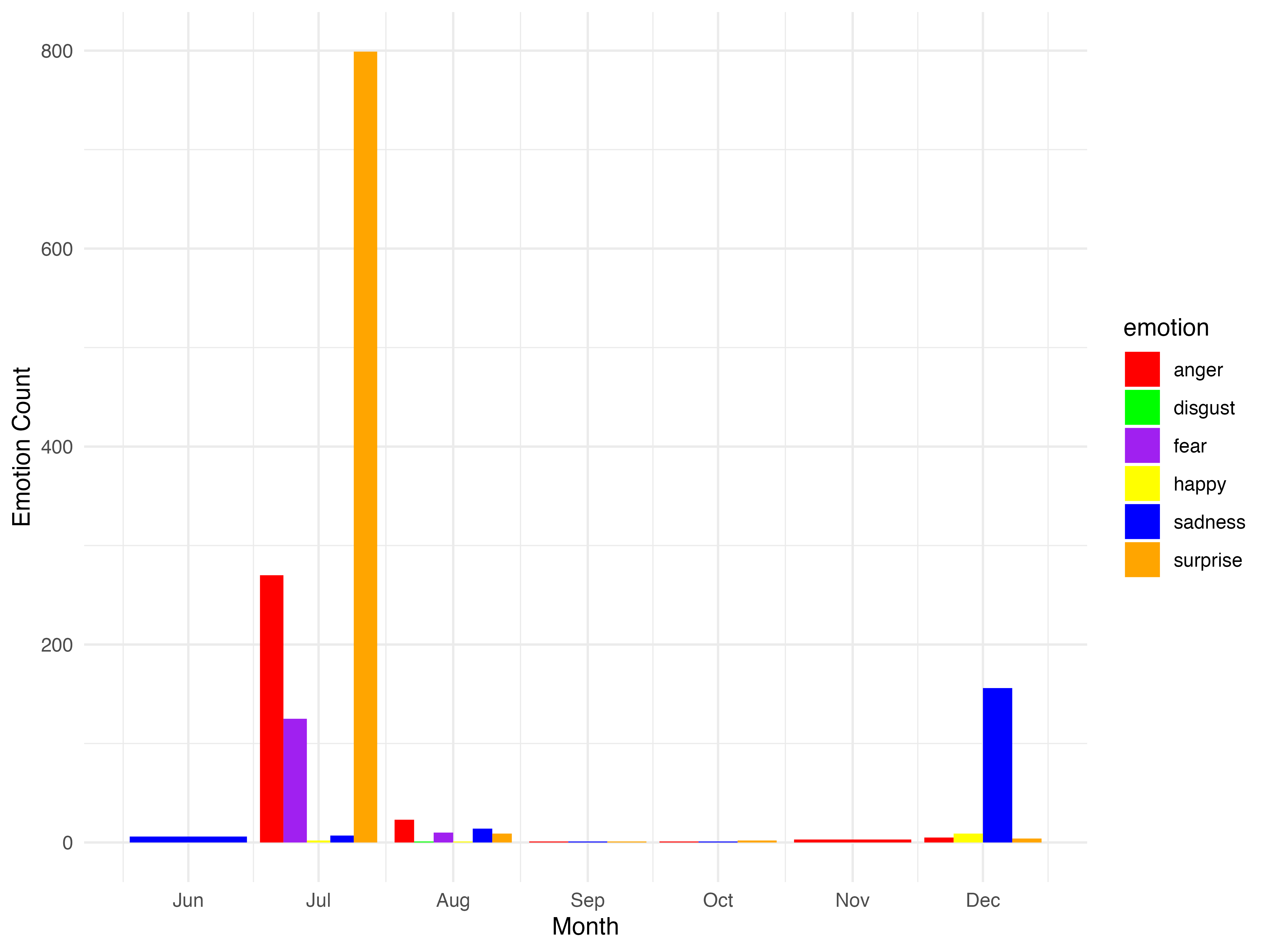} 
  \vspace{1em} 
  \caption{Emotional map throughout 2021.}
  \label{fig:fig8}
\end{figure}

To provide a more detailed year-by-year analysis of emotional trends, distinguishing between 2021 and 2022. In June 2021, sadness was the only discernible emotion, albeit in minimal amounts. By mid-July, all six basic emotions—anger, disgust, fear, happiness, sadness, and surprise—became evident, with surprise, anger, and fear dominating, as shown in Figure 8. Surprise emerged as the most prevalent emotion but declined sharply by mid-August, leaving anger as predominant. Emotions tied to surprise, sadness, and anger persisted in discussions until December 2021, though their overall intensity waned, except for sadness, which notably strengthened by the year’s end.

\begin{figure}[htbp]
  \centering
  \includegraphics[width=0.8\linewidth]{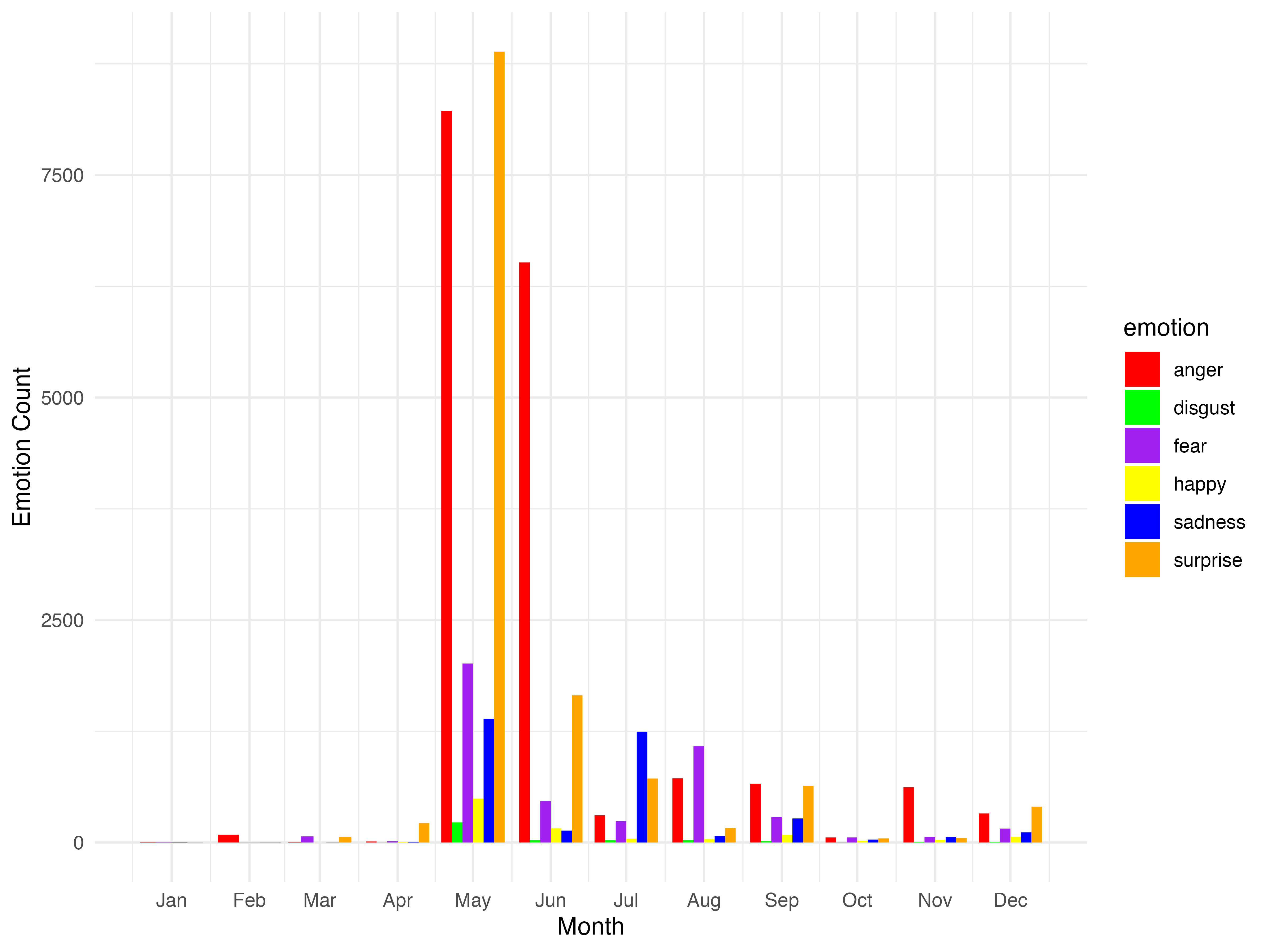} 
  \vspace{1em} 
  \caption{Emotional map throughout 2022.}
  \label{fig:fig9}
\end{figure}

During the first quarter of 2022, spanning January to April, emotional responses to the short film remained relatively subdued. This can be attributed to the limited public discourse on the subject during that time. By April, all six recorded emotional categories exhibited a modest increase, reaching a peak in May, as illustrated in Figure 9. This surge coincided with the highest volume of conversations for the year, with May accounting for 51.9\% of the annual total, as detailed in Table 7.

Subsequently, the volume of conversations declined in June, representing 22.7\% of the year’s discussions. The data collectively indicate that the majority of conversations in 2022 were concentrated in May and June, comprising 76.5\% of the total for the year. In terms of emotional content, the discourse was dominated by three categories: anger (44.5\%), surprise (32.6\%), and fear (11.3\%). These findings are presented in Table 6.

\section{Discussion}
This discussion highlights the vital role of advanced Emotion Recognition Models (ERM) in capturing nuanced Turkish-language public sentiment, particularly during significant societal events. By effectively tracking sentiment shifts, the research demonstrates the model’s practical application to real-world discourse, validating its ability to monitor emotional dynamics over time while shedding light on broader trends in societal polarization.

The fine-tuned BERTurk model, developed with the balanced TREMO dataset, marks a significant advancement in Turkish-language emotion recognition by accurately classifying six basic emotions: anger, disgust, fear, happiness, sadness, and surprise. This study introduces Emotion Recognition Models (ERM) for Turkish, setting a benchmark for emotion classification in the language. Balancing the dataset, with each emotion category capped at 3,003 sentences, was a critical step to address data imbalances and ensure fairness. The model achieved strong results, including 92.47\% accuracy alongside high precision, recall, and F1 scores, underscoring its effectiveness and reliability.

\begin{figure}[htbp]
  \centering
  \includegraphics[width=0.8\linewidth]{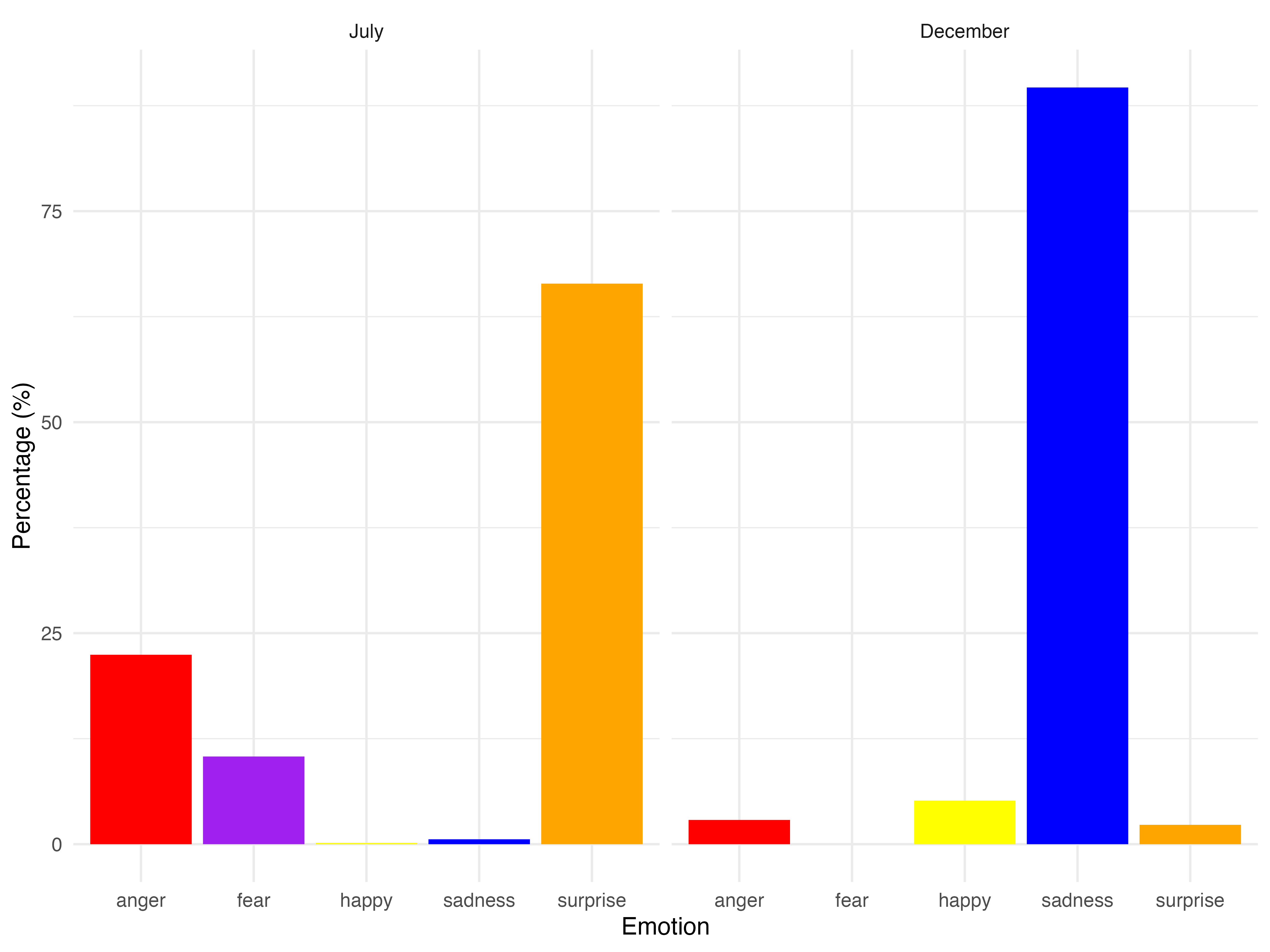} 
  \vspace{1em} 
  \caption{Comparison of emotion distribution percentages for July and December 2021.}
  \label{fig:fig10}
\end{figure}

The model’s application to social media data associated with Sessiz Istila (Silent Invasion) provided valuable insights into public sentiment trends. Analysis revealed that anger and surprise were the predominant emotions, signifying a shift from surprise in 2021 to anger in 2022. In order to gain a deeper understanding of this transition, we will present a detailed analysis of four key events that occurred between 2021 and 2022. As illustrated in Figures 8 and 9, there are two noteworthy occurrences in July and December 2021 and May and June 2022, respectively. The initial findings for 2021, as illustrated in Figure 10, reveal a notable evolution in public sentiment, with key emotional peaks tied to major events. In July, public sentiment was characterized by surprise (66\%) and anger (22\%), driven by reports of illegal immigrants entering Niğde in trucks, which were framed as a ‘silent invasion’. Similarly, by December, sadness became the dominant emotion (90\%), spurred by narratives describing an ‘economic invasion’ tied to the devaluation of the Turkish Lira, where foreign tourists benefitted disproportionately. Together, these events shaped a transition from high-arousal emotions like anger and surprise to a more subdued and reflective state of sadness as socio-economic frustrations mounted over the year.
This temporal evolution, contextualized alongside major events outlined in Figures 8 and 9, underscores the model’s ability to capture dynamic sentiment patterns during pivotal moments of public discourse. The highlighted events substantiate that Sessiz Istila had already emerged as a prominent topic of discussion in 2021, influencing public sentiment across a spectrum of emotions. These findings highlight the interplay between external events, social media narratives, and collective emotional responses, showcasing the role of key incidents in shaping public sentiment over time.

\begin{figure}[htbp]
  \centering
  \includegraphics[width=0.8\linewidth]{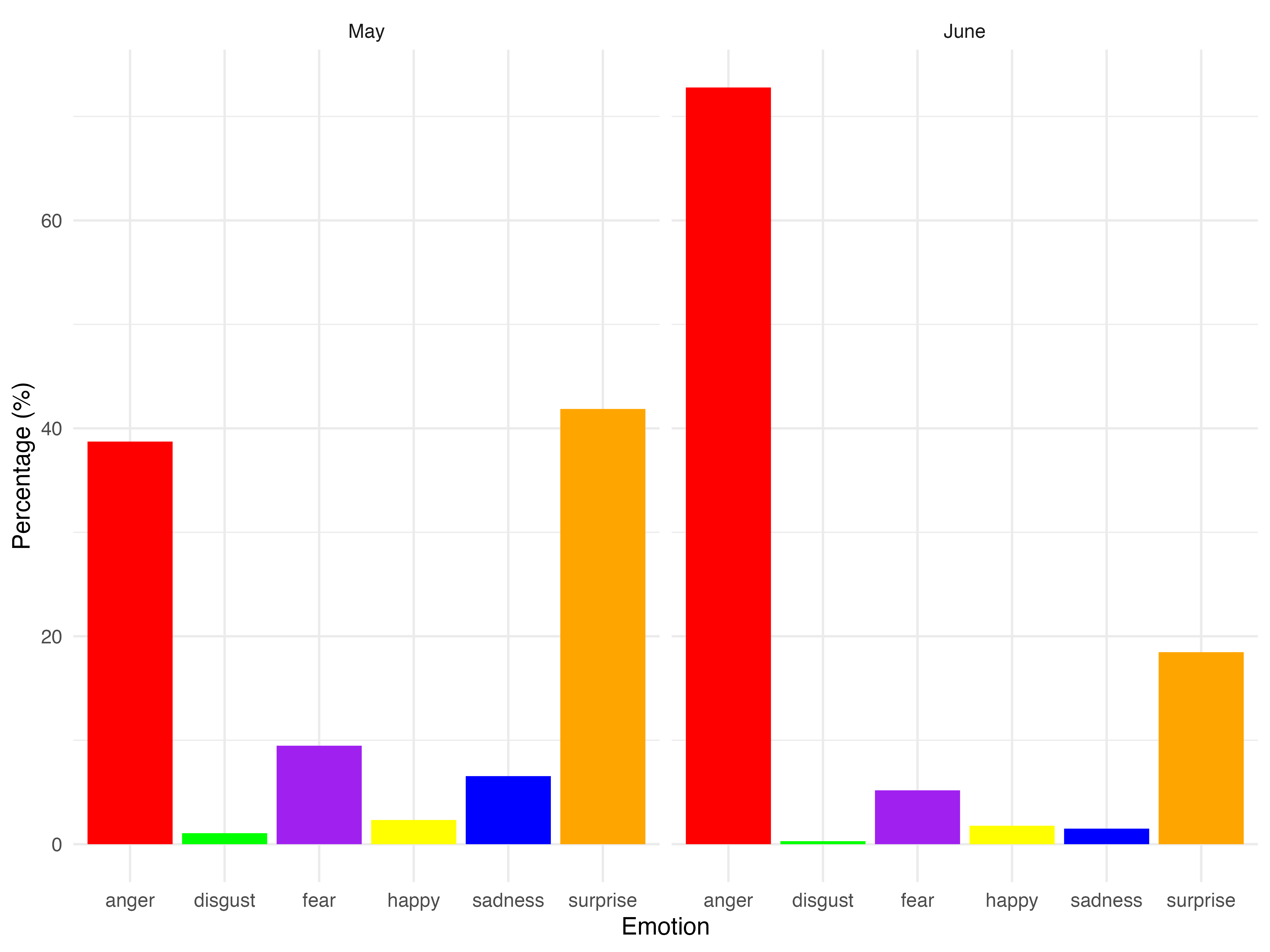} 
  \vspace{1em} 
  \caption{Comparison of emotion distribution percentages for May and June 2022.}
  \label{fig:fig11}
\end{figure}

Emotional intensity peaked in May and June 2022, coinciding with the release of the short movie Sessiz Istila on YouTube on May 3, which drew significant public attention. During this period, anger dominated the emotional landscape, rising sharply from 39\% in May to 73\% in June, reflecting the escalation of negative sentiment as public engagement with the movie intensified. Meanwhile, surprise, initially prominent at 42\% in May, dropped to 18\% in June, indicating a transition from shock to anger as discourse evolved, as shown in Figure 11. Other emotions, such as fear, happiness, sadness, and disgust, remained consistently marginal, collectively accounting for less than 20\% throughout.

These findings are consistent with those of Bahadır Türk (2023)\cite{bahadır2023}, who identified a mix of anxiety, fear, and anger in Turkish political party manifestos, reflecting growing public dissatisfaction with refugees. Similarly, Erbaysal Filibeli \& Öneren Özbek (2023)\cite{erbaysal2023} analyzed YouTube comments on Sessiz Istila, advocating refugee removal and often accompanied by extreme measures or violence. Together, these analyses reveal a broader pattern of heightened emotional intensity, particularly anger and fear, within refugee-related discourse in Türkiye across both online platforms and formal political contexts. The shift from surprise and sadness in 2021 to anger in mid-2022 underscores the increasing polarization and hostility in public discussions, demonstrating the model’s ability to track temporal sentiment trends linked to major societal events. This aligns with earlier research highlighting anger as a significant social signal in online discourse, conveying to others that ‘something unjust is happening here’, as demonstrated in Palmer’s (2024) findings. In this research, surprise plays a transient but critical role in sparking engagement, often setting the stage for sustained reactions like anger.

\subsection{Limitations of the Study}
The study faced several limitations, primarily stemming from the balanced yet size-restricted TREMO dataset, which, while ensuring equal representation of six emotions, constrained the diversity of training samples and potentially impacted the model’s generalizability. Additionally, the model struggled to differentiate between anger and fear due to semantic overlaps in linguistic expressions, highlighting the need for improved contextual representation and advanced attention mechanisms. The simplified six-emotion framework further limited the scope by excluding nuanced emotional categories like anxiety or hope, which could better capture the complexity of human emotions. Lastly, while metrics like accuracy and F1 scores reflected strong performance, they failed to account for the subtleties of emotion recognition, underscoring the importance of incorporating qualitative assessments in future studies.

\subsection{Implications and Further Research}
The insights gained from this study highlight the transformative potential of localized NLP tools, with the developed ERM models offering robust applications for real-time sentiment analysis in Turkish-language contexts. By addressing critical domains including marketing, public relations, and crisis management, these models pave the way for enhanced decision-making through timely and accurate sentiment tracking. Their practical utility underscores the importance of advancing research tailored to regional and linguistic nuances.
To enhance Emotion Recognition Models (ERM), future research should prioritize refining emotion categorization by incorporating nuanced states including anxiety, hope, and satisfaction, creating a more comprehensive sentiment framework. Expanding the underrepresented emotions like happiness and disgust is equally crucial for broadening the analytical scope and better capturing the multifaceted nature of human sentiment. Additionally, employing advanced contextual techniques alongside data augmentation strategies can diversify datasets and improve model generalizability. This approach, as previously outlined by Islam et al. (2024), would enable ERMs to better interpret and respond to complex public sentiment dynamics, particularly when addressing sensitive societal challenges.

\section{Conclusion}
In conclusion, this study makes a notable contribution to Turkish-language emotion recognition by developing and fine-tuning BERTurk-based Emotion Recognition Models (ERM) with high accuracy in classifying six fundamental emotions. Leveraging the TREMO dataset with balanced preprocessing ensured fairness across categories, enabling robust performance in both training and real-world applications, including tracking emotional trends in social media data. For instance, the model effectively captured the shift from surprise to anger during societal events like the release of Sessiz Istila’s movie, highlighting its practical value in understanding public sentiment and societal polarization. However, challenges remain in distinguishing semantically similar emotions and addressing underrepresented categories. Future research should prioritize expanding emotion categories, incorporating advanced contextual techniques, and employing data augmentation to enhance dataset diversity, ultimately improving the model’s generalizability and utility for academic and practical applications.

\bibliographystyle{unsrt}  


\end{document}